\documentclass[twoside,a4paper,11pt]{article}
\usepackage{times}
\usepackage{elcvia}
\usepackage[english]{babel}
\usepackage{float}
\usepackage{graphicx}
\usepackage{adjustbox}

\input{psfig.sty}

\newcommand{\vol}{0}            
\newcommand{\num}{0}            
\newcommand{\begpag}{\thepage}  
\newcommand{\finpag}{7}         
\newcommand{\yearp}{2000}       

\title{BadODD: Bangladeshi Autonomous Driving Object Detection Dataset}

\author{Mirza Nihal Baig$^*$ , Rony Hajong$^*$ , Mahdi Murshed Patwary$^*$ , Mohammad Shahidur Rahman$^*$,\\ Husne Ara Chowdhury$^*$ \\ \\
\centerline{\small \em $^*$ CSE, Shahjalal University of Science and Technology,  Sylhet, Bangladesh} \\
 \\
\centerline{\small }
       }

\begin{document}

\maketitle

\pagestyle{myheadings}

\hrulefill

\begin{abstract}

We propose  a comprehensive dataset for object detection in diverse driving environments across 9 districts in Bangladesh. The dataset, collected exclusively from smartphone cameras, provided a realistic representation of real-world scenarios, including day and night conditions. Most existing datasets lack suitable classes for autonomous navigation on Bangladeshi roads, making it challenging for researchers to develop models that can handle the intricacies of road scenarios. To address this issue, the authors proposed a new set of classes based on characteristics rather than local vehicle names. The dataset aims to encourage the development of models that can handle the unique challenges of Bangladeshi road scenarios for the effective deployment of autonomous vehicles. The dataset did not consist of any online images to simulate real-world conditions faced by autonomous vehicles. The classification of vehicles is challenging because of the diverse range of vehicles on Bangladeshi roads, including those not found elsewhere in the world. The proposed classification system is scalable and can accommodate future vehicles, making it a valuable resource for researchers in the autonomous vehicle sector.

\vspace{0.5cm}
\noindent
{\em Key Words}: Computer Vision, Object Detection, Autonomous Vehicle, Dataset, Vehicle Classification.

\end{abstract}
\hrulefill

\section{Introduction}

Autonomous navigation is advancing towards incredible technology in our daily lives. Many companies integrate autonomous technologies  into their products in various regions and try to achieve level 5 autonomy. Large-scale datasets have contributed significantly to the progress of autonomous navigation. However, in many parts of the world, this technology still has a long way to go. A key challenge is to obtain a diverse set of data to account for extreme corner cases. Most of the algorithms developed for autonomous vehicles are not benchmarked on unstructured and congested data from different parts of the world. In addition, existing datasets do not account for classes that are available on the Indian subcontinent, which is a major disadvantage in advancing autonomous technology.

In this paper, we propose BDOR, a novel dataset for autonomous navigation of Bangladeshi roads that addresses the problems of existing autonomous vehicle datasets. Our dataset consists of 9825 images with 78,943 objects of a bangladesh road under various lightning conditions with 13 classes. As the number of objects per image is higher in Bangladesh roads, this dataset consists of the most common objects that are seen on BD roads.

Furthermore, we propose a new set of classes for classifying vehicles in Bangladesh that are scalable and can be used as a heuristic for autonomous cars. There are a diverse set of vehicles, many of which are manufactured locally in unique shapes and customized to the needs of people. These vehicles are not globally acknowledged by any universal name; rather, local names are provided. This creates a problem for setting class names that are aligned with the existing datasets and account for all of the vehicles. Our proposed classes solved this problem by classifying vehicles based on their characteristics rather than their local names. This set of classes can be scalable to new types of vehicles that may be manufactured locally or imported from other countries.

Another potential use case for this new set of classes is that the characteristics of the vehicle can be used for faster decision-making for autonomous cars. For the existing dataset, each class has the same weight as autonomous vehicles. Our proposed classes have different characteristics; therefore, autonomous vehicles can precompute decisions that can help make accurate decisions in real time. Each characteristic will cater to a different decision-making process for navigation. This will be helpful for creating more scalable dataset in the future.

We provide a detailed analysis of the class distribution in BDOR dataset. While some classes are seen heavily on road such as Person, Autorickshaw, Three Wheeler other classes are rarely seen such as wheel chair, train, construction vehicle. For this reason, the dataset is imbalance for the disproponate ratio of the vehicles seen on road. Furthermore, the unstructure nature of the road scenario and congested objects poses certain problems like occlusion in detecting the object.

Our main contributions are the following:

\begin{itemize}
    \item We released a 2D Object Detection autonomous for autonomous driving on Bangladesh Roads
    \item A new set of scalable classes suitable for classifying bangladeshi vehicles 
    \item We trained and benchmarked existing baseline models and made a comparative analysis of them.
\end{itemize}

\begin{table*}[htbp]
    \centering
    \adjustbox{max width=\textwidth}{
        \begin{tabular}{lccccc}  
            \hline
            \textbf{Feature} & \textbf{SODA10M} & \textbf{BDD100K} & \textbf{Waymo Open Dataset}  \\ \hline  
            Size & 10M (unlabeled) + 20K (labeled) images & 100K video frames & 1.28 TB \\
            Types of Data & Images (front and rear cameras) & Images (various cameras) & LiDAR, Camera, Radar \\
            Object Categories & 6 & 10 & 19 \\
            Scenarios & Diverse weather, time, location & Urban, rural, various weather, day/night & Highway, urban \\
            Annotations & Bounding boxes & Bounding boxes & Bounding boxes, 3D object points \\
            Challenges & Occlusions, scale variations, low resolution & Occlusions, night-time, diverse weather & Proprietary format, limited access \\
            Focus & Self/semi-supervised learning & General 2D object detection & Autonomous driving research \\
            Publicly Available? & Yes & Yes & Partially \\
            \hline
        \end{tabular}
    }
    \caption{Comparison of 2D Object Detection Datasets for Autonomous Vehicles}
    \label{tab:dataset_comparison}
\end{table*}

\section{Dataset}

\subsection{Data Collection}
The data were collected from 9 districts in Bangladesh: Sylhet, Dhaka, Rajshahi, Mymensingh, Maowa, Chittagong, Sirajganj, Sherpur, and Khulna. These cities have different road scenarios and different types of vehicles in various proportions. There are urban and rural areas, highways, and expressway roads with different types of traffic. We collected a dataset from the front of the car. Sometimes, there was car glass in front of the camera, and others time we collected without glass in front. Day and nighttime datasets were collected for different types of roads. To ensure a real-world road scenario, videos were collected while the driver was driving. No online images were collected to ensure  the driving road scenery and quality of the dataset. 

\subsection{Frame Selection}
The method by which we sampled images from video sources was carefully planned to capture the dynamic quality of various urban environments. Because traffic densities vary in different settings, we used a flexible frame-rate sampling strategy to provide a representative and varied dataset.

The frame rate on highways and expressways, which is frequently associated with reduced traffic volumes, was set at one frame per second. This decision was made after a thorough manual review of the film with the goal of eliminating repetition in less dynamic situations while retaining crucial details.

However, we chose a lower frame rate of one frame every two seconds in highly crowded urban regions with increased traffic and pedestrian activity. The necessity of guaranteeing dataset diversity by recording a wider range of circumstances in high-density traffic zones motivated this decision.

The changing frame rates are largely determined by observing the video content. This made it possible for us to customize the sample plan to the unique features of every site, producing a dataset that faithfully captured the dynamic character of both densely populated areas and those with little traffic.

In addition to improving the representativeness of the dataset, this adaptive sampling technique advances our understanding of the traffic patterns in various metropolitan environments. Frame rates were carefully considered in accordance with our dedication to capturing the subtleties of real-world surroundings, providing a strong basis for reliable analysis and model training.

\subsection{Redefining Classes \& Annotation}

Most of the existing object detection datasets have similar classes that do not represent the vehicles on Bangladesh roads. As there are diverse types of vehicles, labelling the vehicles by their local or globally acknowledged names will increase the number of classes, and if new vehicles are manufactured, new classes must be added. Our approach to redefining labels based on the characteristics of vehicles solves this problem.

\begin{table}[H]
    \centering
    \adjustbox{max width=\textwidth}{
        \begin{tabular}{|c|c|c|c|}
            \hline
            \textbf{Class} & \textbf{Wheel Number} & \textbf{Driving Force} & \textbf{Size} \\
            \hline
            Car & 4 wheeler & Diesel, Gas, Fossil Fuel & Medium  \\
            Three Wheeler & 3 Wheeler & Paddle & Small \\
            Autorickshaw & 3 Wheeler & Gas, Electric & Medium \\
            Priority Vehicle & Any vehicle with Siren on Top  & Fossil fuel or paddle & Any size \\
            Bus & 4 Wheeler & Diesel & Big\\
            Truck & 4 or 8 wheeler & fossil fuel or electric & Medium and Small \\
            Cart Car &  3 or 2 Wheeler & Human or Animal Supported (No paddle) & Small \\
            Construction Vehicle & Construction Related Vehicle & Diesel, Fossil Fuel & Medium, Big \\
            Train & Runs on Railway Track & Diesel & Big \\
            Wheelchair & 2 Wheeler with Priority & Human Support & Small \\
            Motorbike & 2 Wheeler & Oil, Electric & Medium \\
            Bicycle & 2 Wheeler & Paddle & Small \\
            \hline
        \end{tabular}
    }
    \caption{Comparison with existing Datasets}
    \label{tab:mytable}
\end{table}

\subsection{Statistical Analysis of Dataset}

In this section, we present a comprehensive statistical analysis of the dataset used for training, validation, and testing our object detection model. The dataset comprised 9,825 images distributed among 13 distinct classes. The dataset was split into training, validation, and testing sets in a 60:20:20 ratio, ensuring that each split included night images.

\textbf{Dataset Overview}

\begin{table}[H]
\centering
\begin{tabular}{|l|c|}
\hline
\textbf{Dataset} & \textbf{Number of Images} \\
\hline
Training Set & 5,896 \\
Validation Set & 1,964 \\
Testing Set & 1,965 \\
\hline
\end{tabular}
\caption{Dataset Overview}
\end{table}

\textbf{Class Distribution}
\begin{table}[H]
\centering
\begin{tabular}{|l|c|c|c|}
\hline
\textbf{Class} & \textbf{Training Set} & \textbf{Validation Set} & \textbf{Testing Set} \\
\hline
Person & 18,010 (38.22\%) & 6,202 (38.69\%) & 6,104 (38.64\%) \\
Three-Wheeler & 5,710 (12.12\%) & 1,929 (12.04\%) & 1,961 (12.41\%) \\
Motorbike & 3,749 (7.96\%) & 1,218 (7.60\%) & 1,254 (7.94\%) \\
Auto Rickshaw & 10,614 (22.53\%) & 3,638 (22.70\%) & 3,489 (22.09\%) \\
Car & 3,785 (8.03\%) & 1,305 (8.14\%) & 1,233 (7.81\%) \\
Truck & 2,296 (4.87\%) & 777 (4.85\%) & 782 (4.95\%) \\
Bus & 1,885 (4.00\%) & 589 (3.67\%) & 613 (3.88\%) \\
Bicycle & 673 (1.43\%) & 256 (1.60\%) & 241 (1.53\%) \\
Priority Vehicle & 229 (0.49\%) & 70 (0.44\%) & 64 (0.41\%) \\
Cart Vehicle & 141 (0.30\%) & 38 (0.24\%) & 44 (0.28\%) \\
Construction Vehicle & 23 (0.05\%) & 6 (0.04\%) & 11 (0.07\%) \\
Wheelchair & 2 (0.00\%) & 0 (0.00\%) & 1 (0.01\%) \\
Train & 1 (0.00\%) & 0 (0.00\%) & 0 (0.00\%) \\
\hline
\end{tabular}
\caption{Total Class Distribution in Each Split}
\end{table}

\begin{figure}[p] 
    \centering
    \includegraphics[width=\linewidth]{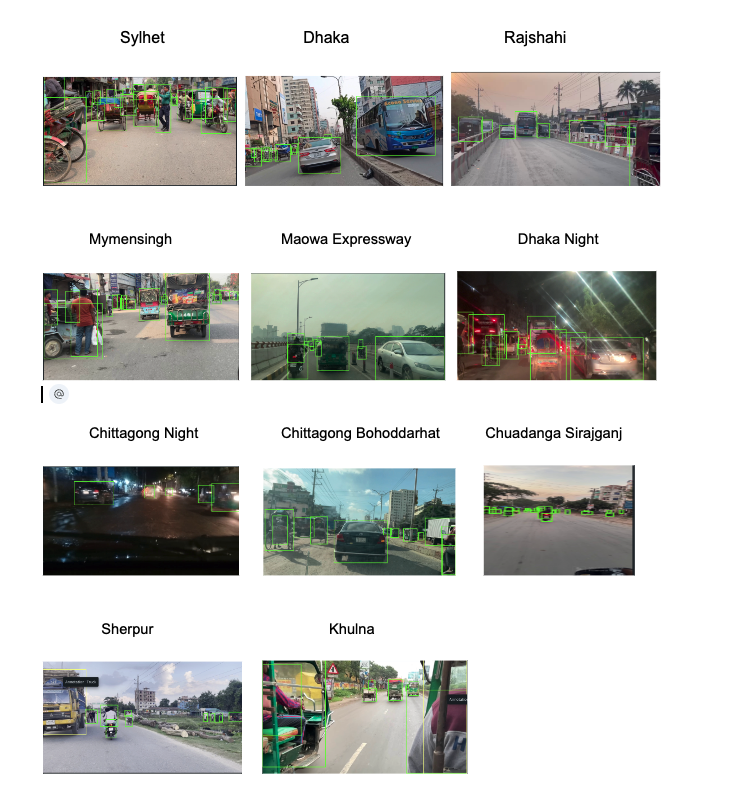} 
    \caption{Sample Dataset}
    \label{fig:fullpage}
\end{figure}

\section{Model}

In this section, we elucidate the intricacies of the training regimen and architectural configurations for YOLOv5 and YOLOv8. Our primary objective is to draw a comparative analysis of their hyperparameters and performance, as gauged by the mean Average Precision (mAP) metric on a designated object detection dataset.

\subsection{YOLOv5}

The YOLOv5 model stands as a vanguard in the realm of object detection, esteemed for its swift processing capabilities and commendable accuracy. Its architecture is founded upon a deep neural network, striking a judicious equilibrium between speed and efficacy.

\subsubsection{Hyperparameters}

Our training process for the YOLOv5 model adhered to a set of meticulously chosen hyperparameters, as delineated in Table \ref{tab:yolov5_hyperparameters}.

\begin{table}[h]
    \centering
    \begin{tabular}{|c|c|}
        \hline
        \textbf{Hyperparameter} & \textbf{Value} \\
        \hline
        Batch Size & 64 \\
        Learning Rate & 0.001 \\
        Number of Epochs & 10 \\
        Optimizer & Adam \\
        Weight Decay & $1 \times 10^{-4}$ \\
        Input Image Size & 416x416 \\
        \hline
    \end{tabular}
    \caption{Hyperparameters employed for training YOLOv5.}
    \label{tab:yolov5_hyperparameters}
\end{table}

\subsubsection{Results}

Upon completion of our training regimen, the YOLOv5 model manifested a commendable mAP score of 0.6 when evaluated on our designated object detection dataset.

\subsection{YOLOv8}

\subsubsection{Hyperparameters}

\begin{table}[H]
    \centering
    \begin{tabular}{|c|c|}
        \hline
        \textbf{Hyperparameter} & \textbf{Value} \\
        \hline
        Batch Size & 64 \\
        Learning Rate & 0.001 \\
        Number of Epochs & 10 \\
        Optimizer & Adam \\
        Weight Decay & $1 \times 10^{-4}$ \\
        Input Image Size & 416x416 \\
        \hline
    \end{tabular}
    \caption{Hyperparameters employed for training YOLOv8.}
    \label{tab:yolov5_hyperparameters}
\end{table}

\subsubsection{Results}
Upon completion of our training regimen, the YOLOv8 model manifested a commendable mAP score of 0.7 when evaluated on our designated object detection dataset.

\subsection{Comparison}

We present a juxtaposition of the hyperparameters and mAP results for YOLOv5 and YOLOv8 in Table \ref{tab:yolov_comparison}.

\begin{table}[h]
    \centering
    \begin{tabular}{|c|c|c|}
        \hline
        \textbf{Model} & \textbf{mAP} & \textbf{Training Time} \\
        \hline
        YOLOv5 & 0.6 & 1 hour \\
        YOLOv8 & 0.7 & 1 hour \\
        \hline
    \end{tabular}
    \caption{Comparative analysis of YOLOv5 and YOLOv8.}
    \label{tab:yolov_comparison}
\end{table}

\section{Conclusion}

The research discusses a unique and complex task of object detection in diverse driving environments across 9 districts in Bangladesh. The dataset, collected exclusively from smartphone cameras, provides a comprehensive representation of real-world scenarios, encompassing both day and night conditions. The limitations of existing datasets available online don't have vehicles that are available on Bangladesh roads or suitable for autonomous navigation in Bangladesh. The challenges in object detection require robust algorithms capable of identifying and localizing objects in dynamic and challenging environments, such as varying lighting conditions, different road types, and the unique characteristics of each district. The dataset was collected using smartphone cameras to ensure authenticity and to simulate real-world conditions faced by autonomous vehicles. The classification of vehicles was challenging due to the many types of vehicles that cannot be seen elsewhere in the world and have local vehicle names. A new set of classes based on characteristics rather than local vehicle names was proposed to overcome this problem. The dataset will set a new benchmark for developing better models on Bangladesh roads and will open up many research opportunities for the autonomous vehicle sector.

\vspace{10pt}

Key highlights of our work:

\begin{itemize}
    \item Object detection in diverse Bangladesh driving environments
    \item Exclusive dataset from smartphone cameras for real-world scenarios
    \item 9 districts covered, including day and night conditions
    \item Limitations of existing datasets addressed
    \item Challenges: robust algorithms for dynamic environments
    \item Data collection: smartphone cameras for authenticity
    \item New class system for scalable vehicle classification
    \item Benchmark for better models on Bangladesh roads
    \item Research opportunities for the autonomous vehicle sector.
\end{itemize}

\end{document}